\documentclass[acmtog]{acmart}

\usepackage{booktabs}
\usepackage[ruled]{algorithm2e}

\SetAlFnt{\small}
\SetAlCapFnt{\small}
\SetAlCapNameFnt{\small}
\SetAlCapHSkip{0pt}

\copyrightyear{2026}
\acmYear{2026}
\ifdefined\setcctype
  \setcopyright{cc}
  \setcctype{by}
\else
  \setcopyright{none}
\fi
\acmConference[SA Conference Papers '26]{SIGGRAPH Asia 2026 Conference Papers}{December 01--04, 2026}{Kuala Lumpur, Malaysia}
\acmBooktitle{SIGGRAPH Asia 2026 Conference Papers (SA Conference Papers '26), December 01--04, 2026, Kuala Lumpur, Malaysia}
\acmDOI{10.1145/3829340.3842368}
\acmISBN{979-8-4007-2842-6/2026/12}

\usepackage{url}
\usepackage{multirow}
\usepackage{graphicx}
\usepackage{colortbl,xcolor}
\usepackage{soul}
\colorlet{llgray}{lightgray!40}
\sethlcolor{llgray}
\usepackage{mathtools}
\usepackage{caption}
\usepackage{physics}
\usepackage{amsmath}
\usepackage{bm}

\newcommand{\vx}{\mathbf{x}}
\newcommand{\vc}{\mathbf{c}}

\begin{document}

\title{DiffusionOPD: A Unified Perspective of On-Policy Distillation in Diffusion Models}

\author{Quanhao Li}
\authornote{These authors contributed equally.}
\email{liqh24@m.fudan.edu.cn}
\affiliation{
  \institution{Fudan University}
  \city{Shanghai}
  \country{China}
}

\author{Junqiu Yu}
\authornotemark[1]
\email{michaelyu1101@163.com}
\affiliation{
  \institution{Fudan University}
  \city{Shanghai}
  \country{China}
}

\author{Kaixun Jiang}
\email{kxjiang22@m.fudan.edu.cn}
\affiliation{
  \institution{Fudan University}
  \city{Shanghai}
  \country{China}
}

\author{Yujie Wei}
\email{yjwei22@m.fudan.edu.cn}
\affiliation{
  \institution{Fudan University}
  \city{Shanghai}
  \country{China}
}

\author{Zhen Xing}
\email{xingzhen.xz@alibaba-inc.com}
\affiliation{
  \institution{Alibaba Group}
  \city{Hangzhou}
  \country{China}
}

\author{Pandeng Li}
\email{lipandeng.lpd@alibaba-inc.com}
\affiliation{
  \institution{Alibaba Group}
  \city{Hangzhou}
  \country{China}
}

\author{Ruihang Chu}
\email{churuihang.crh@alibaba-inc.com}
\affiliation{
  \institution{Alibaba Group}
  \city{Hangzhou}
  \country{China}
}

\author{Shiwei Zhang}
\email{zhangjin.zsw@alibaba-inc.com}
\affiliation{
  \institution{Alibaba Group}
  \city{Hangzhou}
  \country{China}
}

\author{Yu Liu}
\email{ly103369@alibaba-inc.com}
\affiliation{
  \institution{Alibaba Group}
  \city{Hangzhou}
  \country{China}
}

\author{Zuxuan Wu}
\authornote{Corresponding author.}
\email{zxwu@fudan.edu.cn}
\affiliation{
  \institution{Fudan University}
  \city{Shanghai}
  \country{China}
}

\renewcommand{\shortauthors}{Li et al.}

\begin{abstract}
Reinforcement learning has emerged as a powerful tool for improving diffusion-based text-to-image models, but existing methods are largely limited to single-task optimization. Extending RL to multiple tasks is challenging: joint optimization suffers from cross-task interference and imbalance, while cascade RL is cumbersome and prone to catastrophic forgetting.
We propose \textit{DiffusionOPD}, a new multi-task training paradigm for diffusion models based on Online Policy Distillation (OPD). DiffusionOPD first trains task-specific teachers independently, then distills their capabilities into a unified student along the student’s own rollout trajectories. This decouples single-task exploration from multi-task integration and avoids the optimization burden of solving all tasks jointly from scratch.
Theoretically, we lift the OPD framework from discrete tokens to continuous-state Markov processes, deriving a \emph{closed-form} per-step KL objective that unifies both stochastic SDE and deterministic ODE refinement via mean-matching. We formally and empirically demonstrate that this analytic gradient provides lower variance and better generality compared to conventional PPO-style policy gradients.
Extensive experiments show that DiffusionOPD consistently surpasses both multi-reward RL and cascade RL baselines in training efficiency and final performance, while achieving state-of-the-art results on all evaluated benchmark including DrawBench, OCR, and GenEval.
\end{abstract}



\begin{teaserfigure}
    \centering
    \includegraphics[width=\textwidth]{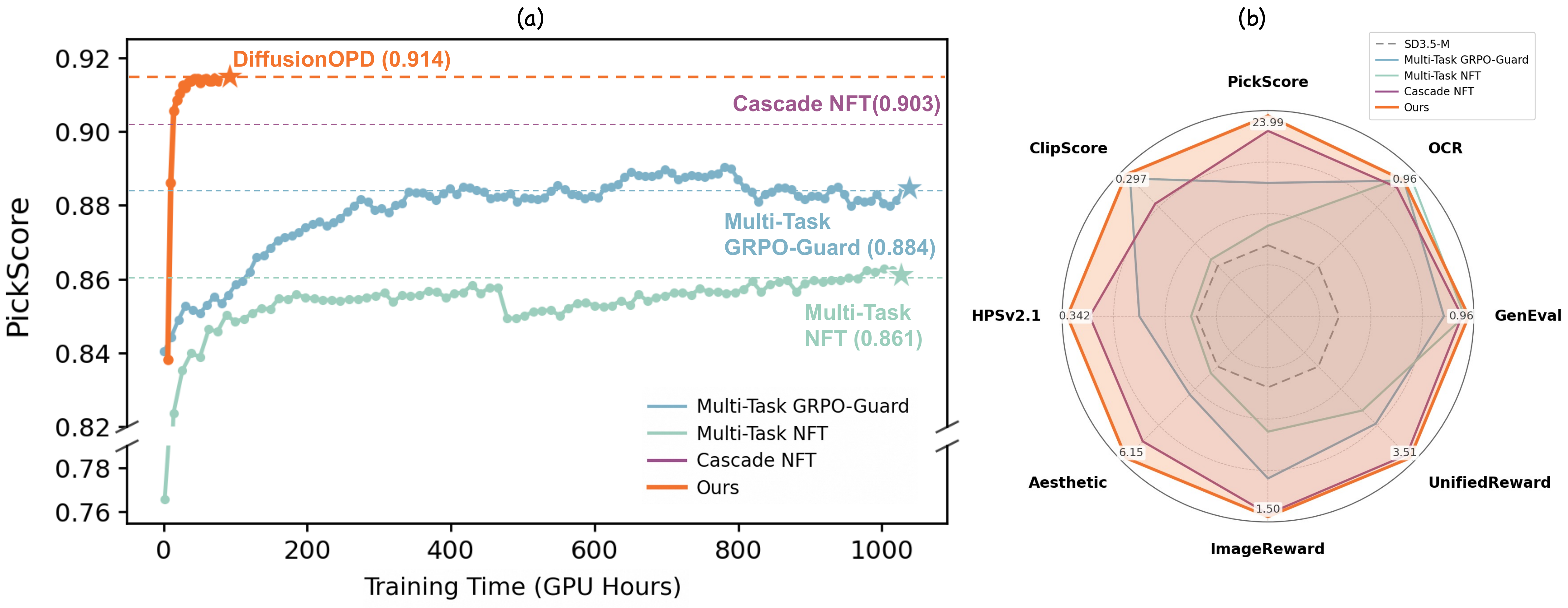}
    \caption{(a) DiffusionOPD exhibits significantly faster convergence and a higher performance ceiling than all multi-task reinforcement learning baselines. (b) DiffusionOPD consistently outperforms all baselines across multiple domains, including GenEval, OCR, and aesthetics.}
    \label{fig:teaser}
\end{teaserfigure}

\maketitle

\section{Introduction}
Reinforcement learning (RL)~\cite{schulman2017proximal, shao2024deepseekmath, rafailov2023direct} has recently emerged as a powerful paradigm for improving diffusion-based models~\cite{rombach2022high, labs2025flux1kontextflowmatching, mao2026wan, li2026flashmotion, Li_2025_ICCV, xing2024simda, miao2026rose, zuo2026filmworld, wei2024dreamvideo, wei2025dreamrelation, xing2023vidiff, xing2024survey, xing2024aid, wei2026routing}. A growing body of work~\cite{zheng2025diffusionnft, liu2026flow, wang2025grpo, wallace2024diffusion, xu2023imagereward, xue2025dancegrpo} has shown that RL can substantially boost performance when optimizing against a single reward
signal. 
However, these gains are typically task-specific. In practice, users often expect a single model to satisfy multiple objectives simultaneously, for example, generating images that are both aesthetically pleasing and faithful to textual instructions. This mismatch between single-objective optimization and multi-objective user demand naturally motivates the study of multi-task RL.

Multi-task RL aims to equip a single diffusion model with multiple capabilities by optimizing it over several task-specific rewards. Existing approaches mainly follow two paradigms. The first is joint optimization, which trains all tasks simultaneously within a unified framework. Although appealing in principle, this strategy often suffers from two fundamental challenges: objective conflict across tasks and task-difficulty imbalance. Different tasks may induce inconsistent optimization directions, causing cross-task interference during training, while easier tasks tend to dominate the learning dynamics and suppress signals from more challenging ones.

The second paradigm is cascade RL \cite{zheng2025diffusionnft, mao2026wan}, which optimizes the policy on different tasks sequentially rather than simultaneously, avoiding direct gradient conflict within each training stage. However, this strategy is often cumbersome in practice, as it requires multiple training stages, carefully designed schedules, and task-specific hyperparameter. It is also prone to catastrophic forgetting~\cite{kirkpatrick2017overcoming}, where adaptation to later tasks can degrade performance on those learned earlier.

To address the reward conflict in joint optimization and the cumbersome training procedure of cascade optimization, we argue that multi-task RL should be decoupled into two distinct processes:\textbf{ single-task on-policy exploration} and \textbf{multi-task capability integration}. Motivated by the success of On-Policy Distillation (OPD)~\cite{thinkingmachines2026opd}, we propose \textbf{DiffusionOPD}, an on-policy distillation framework for diffusion models. Concretely, we first train a set of task-specific teacher models, each optimized independently for a single task, and then distill their capabilities into a unified student model. This avoids cross-task interference during teacher training and eliminates the student's exploration burden to solve all tasks from scratch.

To extend OPD from LLMs to diffusion models, we first derive a diffusion-domain OPD objective. Specifically, we lift the original formulation from autoregressive token transitions to continuous-state denoising transitions, and model the diffusion denoising process as a discrete-time Markov chain induced by the reverse-time SDE~\cite{liu2026flow}. Under this view, both the student and the teacher define one-step Gaussian transition kernels at each denoising state. Because these kernels share the same covariance, their reverse KL admits a closed-form expression, yielding an analytic OPD objective for diffusion models.

Given this objective, a straightforward choice is to follow~\cite{thinkingmachines2026opd} and optimize the student with a PPO-style objective, using the per-step reverse KL as a dense reward and treating the teacher as a process-level reward model~\cite{wang2024math, zhang2025lessons, cui2025process} along the student trajectory. However, our derivation reveals that this formulation introduces an additional score-function term proportional to Gaussian noise. Although unbiased in expectation, this term increases gradient variance, making PPO~\cite{schulman2017proximal} an unnecessarily noisy way to optimize a quantity that is already available in closed form.

We therefore directly optimize the closed-form KL objective rather than relying on a PPO-style surrogate. This design reduces gradient variance and yields stronger empirical performance. Moreover, it naturally extends to deterministic ODE samplers, where it recovers direct transition matching, thereby offering a unified view of on-policy distillation across different diffusion samplers.

More importantly, our framework is not limited to the closed-form reverse-KL objective derived above. Once the student generates on-policy rollouts, the teacher can supervise the visited denoising states using a broad family of existing distillation objectives~\cite{yin2024improved, yin2024onestep, luo2025learning}. DiffusionOPD should therefore be viewed not merely as a reverse-KL method, but more generally as a unified framework for on-policy distillation in diffusion models.

We further evaluate \textbf{DiffusionOPD} in the multi-task setting, where it consistently surpasses all multi-task RL baselines across diverse benchmarks in both training efficiency and final performance. We also conduct ablations on key design choices, including the distillation objective, loss formulation, and sampler noise level.

Our contributions can be summarized as follows:
\begin{itemize}
    \item We propose \textbf{DiffusionOPD}, a new on-policy distillation paradigm for multi-task training of diffusion models, where domain-specific teachers supervise a unified student along its own rollout trajectories.

    \item We establish a principled framework for on-policy diffusion distillation by deriving a unified closed-form KL objective for both stochastic and deterministic samplers, enabling lower-variance optimization than PPO-style policy gradients.

    \item We validate DiffusionOPD through multi-task experiments and ablations, showing consistent gains over prior baselines in both training efficiency and final performance, with state-of-the-art results on aesthetics, OCR, and GenEval. Our ablations further highlight the impact of key design choices.

\end{itemize}

\section{Related Works}
\subsection{RL for Diffusion.}
Reinforcement learning (RL) has recently emerged as an effective paradigm for improving diffusion-based text-to-image models~\cite{rombach2022high}. Building on advances in Reinforcement Learning~\cite{schulman2017proximal, shao2024deepseekmath, rafailov2023direct}, a growing line of work has adapted RL to diffusion generation and shown that it can substantially improve model behavior under task-specific reward signals, such as aesthetic quality, text rendering accuracy, and compositional alignment~\cite{zheng2025diffusionnft, liu2026flow, wang2025grpo, wallace2024diffusion, xu2023imagereward, xue2025dancegrpo}. 
Most existing methods, however, focus on optimizing a single reward at a time, yielding task-specialized improvements rather than a unified model that performs well across multiple objectives. In practice, users often expect a single text-to-image model to satisfy several desiderata simultaneously, such as visual appeal, prompt faithfulness, and OCR correctness. This gap has motivated growing interest in extending RL for diffusion models from single-task optimization to the multi-task setting.

\subsection{Diffusion Distillation}
Diffusion distillation aims to transfer the knowledge of a teacher diffusion model to a student model. Most prior work in this area has focused on \emph{step distillation}, where a many-step teacher is compressed into a few-step student for more efficient inference. Existing approaches can be broadly grouped into two categories. 
\textit{Trajectory distillation}~\cite{on_distill,salimans2021progressive,song2023consistency,song2024improved,luo2023latent} distills the teacher’s denoising process by imitating intermediate transitions or enforcing consistency across timesteps. 
\textit{Distribution matching} methods, on the other hand, train student models by aligning their distributions with those of the teacher at selected timesteps, including Diffusion-GAN hybrids~\cite{sauer2023adversarial, ufogen} and score-distillation methods~\cite{wang2023prolificdreamer, sid, yin2023one, yin2024improved, luo2025learning}. 
In contrast to this line of work, we do not use distillation for step reduction. Instead, we study how to distill multiple reward-specialized teachers into a single aligned student in the multi-task setting, using task-specific teachers to provide dense supervision for capability integration.
\section{Method}
\label{sec:method}
\subsection{Preliminary: OPD in the LLM Domain}
\label{sec:llm-opd}

Let $\pi_\theta$ denote the student language model and let $\pi^\star$ denote a
frozen teacher. For a token sequence $x=(x_1,\dots,x_T)$, both policies
factorize autoregressively:
\begin{equation}
\label{eq:llm-factorization}
\pi_\theta(x) \;=\; \prod_{t=1}^{T} \pi_\theta(x_t \mid x_{<t}),
\qquad
\pi^\star(x) \;=\; \prod_{t=1}^{T} \pi^\star(x_t \mid x_{<t}).
\end{equation}

On-policy distillation~\cite{thinkingmachines2026opd} lets the student
autoregressively generate a full sequence from its own policy, and then trains
the student to match the teacher on the prefixes that the student itself
visits. A natural sequence-level objective is therefore the reverse-KL under
student-generated trajectories:
\begin{equation}
\label{eq:llm-opd-seq}
\mathcal{L}_{\mathrm{OPD}}^{\mathrm{LLM}}(\theta)
\;=\;
\mathrm{KL}\!\bigl(\pi_\theta(\cdot)\,\Vert\,\pi^\star(\cdot)\bigr)
\;=\;
\mathbb{E}_{x \sim \pi_\theta}
\!\left[
\log \frac{\pi_\theta(x)}{\pi^\star(x)}
\right].
\end{equation}
where the expectation is taken over full sequences sampled from the student model $\pi_\theta$.
Using the autoregressive factorization, the sequence-level KL decomposes
exactly into a sum of per-step conditional KLs evaluated along the
student's own trajectory:
\begin{equation}
\label{eq:llm-opd-stepwise}
\mathcal{L}_{\text{OPD}}^{\text{LLM}}(\theta)
\;=\;
\mathbb{E}_{x \sim \pi_\theta}\!\left[
  \sum_{t=1}^{T}
  \mathrm{KL}\!\Bigl(
     \pi_\theta(\cdot \mid x_{<t})
     \,\big\Vert\,
     \pi^\star(\cdot \mid x_{<t})
  \Bigr)
\right].
\end{equation}
For LLMs, this inner KL is a discrete distribution over a finite vocabulary $\mathcal V$,
so it admits a \emph{closed form} as shown below.
\[
\mathrm{KL}\!\Bigl(
\pi_\theta(\cdot \mid x_{<t})
\,\big\Vert\,
\pi^\star(\cdot \mid x_{<t})
\Bigr)
=
\sum_{v\in\mathcal V}
\pi_\theta(v\mid x_{<t})
\log
\frac{\pi_\theta(v\mid x_{<t})}{\pi^\star(v\mid x_{<t})}.
\]
In contrast to standard on-policy reinforcement learning, where the model generates a full response and receives only an outcome-level scalar
reward, OPD provides token-level \emph{dense} supervision. The student receives a full next-token distributional target from the teacher at every decoding step along its own trajectory. This allows the objective to be optimized as an analytic per-step KL via direct backpropagation, avoiding the high-variance policy gradients inherent in sparse reward settings.

\subsection{DiffusionOPD}
\subsubsection{Lifting OPD to a Continuous-State Markov Chain}
We reinterpret \eqref{eq:llm-opd-stepwise} as a statement about
\emph{any} discrete-time Markov chain in which the student and teacher
share the same state space and transition kernel structure. Concretely,
let $x_{t_0}, x_{t_1}, \dots, x_{t_N}$ be a trajectory of states and let
$p_S(\cdot \mid x_{t_j})$ and $p_T(\cdot \mid x_{t_j})$ denote the student
and teacher \emph{one-step} transition kernels. Replacing
``$\pi_\theta(\cdot \mid x_{<t})$'' by ``$p_S(\cdot \mid x_{t_j})$'' and
analogously for $\pi^\star$, the OPD objective becomes
\begin{equation}
\label{eq:opd-generic}
\mathcal{L}_{\text{OPD}}(\theta)
\;=\;
\mathbb{E}_{x_{0:N} \sim p_S}\!\left[
  \sum_{j=0}^{N-1}
  \mathrm{KL}\!\Bigl(
     p_S(\cdot \mid x_{t_j})
     \,\big\Vert\,
     p_T(\cdot \mid x_{t_j})
  \Bigr)
\right].
\end{equation}
Two structural properties of \eqref{eq:opd-generic} survive the lift:
\textbf{(i)} the trajectory is sampled from the student (on-policy), and
\textbf{(ii)} the per-step KL must be available in closed form so we never
need the REINFORCE trick.

\subsubsection{Per-Step Gaussian Transitions}
For a flow-matching model on latents $x \in \mathbb{R}^d$, we follow
Flow-GRPO~\cite{liu2026flow} and discretize the reverse-time SDE by Euler--Maruyama on a
schedule $1 = t_0 > t_1 > \cdots > t_N = 0$ with step size
$\Delta t_j := t_{j+1} - t_j < 0$. Let
$\sigma_t \;=\; a\,\sqrt{t/(1-t)}$ denote the SDE diffusion coefficient, where $a$ is the global noise
level. Writing $v_j^{S} := v_\theta(x_{t_j}, t_j)$ for the student
velocity, the student SDE step is
\begin{equation}
\label{eq:sde-step}
\begin{aligned}
x_{t_{j+1}}
&= x_{t_j}
   + \Bigl[\,v_j^{S} + \tfrac{\sigma_{t_j}^{2}}{2\,t_j}\bigl(x_{t_j} + (1{-}t_j)\,v_j^{S}\bigr)
    \Bigr]\Delta t_j \quad + \sigma_{t_j}\sqrt{-\Delta t_j}\;\varepsilon_j
\end{aligned}
\end{equation}
where $\varepsilon_j \sim \mathcal{N}(0, I_d)$ injects stochasticity.
Collecting the deterministic part of \eqref{eq:sde-step} and abbreviating
the per-step variance as $\bar\sigma_j^{\,2} \;:=\; \sigma_{t_j}^{2}\,(-\Delta t_j)$, the one-step transition kernel is the Gaussian
\begin{equation}
\label{eq:student-kernel}
p_S\bigl(x_{t_{j+1}}\,\big|\,x_{t_j}\bigr)
\;=\;
\mathcal{N}\!\bigl(\mu_S(x_{t_j}),\,\bar\sigma_j^{\,2}\,I_d\bigr),
\end{equation}
with student transition mean
\begin{equation}
\label{eq:student-mean}
\begin{aligned}
\mu_S(x_{t_j})
&= \Bigl(1 + \tfrac{\sigma_{t_j}^{2}}{2\,t_j}\,\Delta t_j\Bigr)\,x_{t_j} + \Bigl(1 + \tfrac{\sigma_{t_j}^{2}(1-t_j)}{2\,t_j}\Bigr)\,v_j^{S}\,\Delta t_j.
\end{aligned}
\end{equation}
We thus construct the teacher kernel $p_T$ by the \emph{same} formulas
\eqref{eq:student-kernel}--\eqref{eq:student-mean} on the \emph{same}
scheduler and noise level, with the student velocity replaced by the
frozen teacher velocity $v_j^{T} := v_\phi(x_{t_j}, t_j)$:
\begin{equation}
\label{eq:teacher-kernel}
p_T\bigl(x_{t_{j+1}}\,\big|\,x_{t_j}\bigr)
\;=\;
\mathcal{N}\!\bigl(\mu_T(x_{t_j}),\,\bar\sigma_j^{\,2}\,I_d\bigr),
\end{equation}
\begin{equation}
\label{eq:teacher-mean}
\begin{aligned}
\mu_T(x_{t_j})
&= \Bigl(1 + \tfrac{\sigma_{t_j}^{2}}{2\,t_j}\,\Delta t_j\Bigr)\,x_{t_j} + \Bigl(1 + \tfrac{\sigma_{t_j}^{2}(1-t_j)}{2\,t_j}\Bigr)\,v_j^{T}\,\Delta t_j.
\end{aligned}
\end{equation}

\subsubsection{Closed-Form Reverse KL between Same-Covariance Gaussians}
Since the per-step covariance $\bar\sigma_j^{\,2} I_d$ depends only on the
scheduler $(t_j,\Delta t_j)$ and the global noise level $a$, it is identical
for the student and teacher. Moreover, under on-policy distillation, both
transition kernels are evaluated at the same student-rollout state $x_{t_j}$.
Therefore, $p_S$ and $p_T$ differ only in their means,
$\mu_S(x_{t_j})$ and $\mu_T(x_{t_j})$, while sharing the same covariance.
For two $d$-dimensional Gaussians with common covariance $\Sigma$,
\[
\mathrm{KL}\!\bigl(
  \mathcal{N}(\mu_1,\Sigma)\,\Vert\,\mathcal{N}(\mu_2,\Sigma)
\bigr)
\;=\;
\tfrac{1}{2}\,(\mu_1 - \mu_2)^{\!\top} \Sigma^{-1}(\mu_1 - \mu_2).
\]
Specializing to $\Sigma = \sigma_j^2 I_d$ gives
\begin{equation}
\label{eq:gauss-kl-iso}
\mathrm{KL}\!\bigl(
  \mathcal{N}(\mu_1,\sigma_j^2 I)\,\Vert\,\mathcal{N}(\mu_2,\sigma_j^2 I)
\bigr)
\;=\;
\frac{\|\mu_1-\mu_2\|_2^2}{2\sigma_j^2}.
\end{equation}
This expression is exact and introduces no Monte-Carlo variance, since the
sample noise $\varepsilon_j$ cancels analytically.

Plugging \eqref{eq:student-kernel}--\eqref{eq:teacher-mean} and
Eq.~\eqref{eq:gauss-kl-iso} into the generic OPD objective
\eqref{eq:opd-generic} yields
\begin{equation}
\label{eq:opd-diffusion}
\mathcal{L}_{\text{OPD}}^{\text{diffusion}}(\theta)
\;=\;
\mathbb{E}_{x_{0:N} \sim p_{S,\theta}}\!\left[
  \sum_{j=0}^{N-1}
  \frac{
    \bigl\|
      \mu_S(x_{t_j}; \theta) - \mu_T(x_{t_j})
    \bigr\|_2^{2}
  }{2\,\sigma_j^{2}}
\right].
\end{equation}

\subsubsection{Deterministic Regime: Direct $L_2$ Matching}
In the LLM setting, reverse KL is the natural OPD objective because the model
defines a stochastic next-token distribution at each prefix, so matching the
teacher necessarily amounts to matching conditional distributions.
By contrast, under the deterministic ODE Euler update in diffusion models, the
next state is uniquely determined by the current latent $x_{t_j}$.
For a given $x_{t_j}$, the student and teacher therefore induce two
deterministic transition targets, $\mu_S(x_{t_j};\theta)$ and
$\mu_T(x_{t_j})$, respectively.
In this regime, we use a direct squared $L_2$ transition-matching surrogate:
\begin{equation}
\label{eq:opd-ode}
\boxed{
\mathcal{L}_{\text{OPD}}^{\text{diffusion-ODE}}(\theta)
\;=\;
\mathbb{E}_{x_{0:N}\sim p_{S,\theta}}\!\left[
  \sum_{j=0}^{N-1}
  \tfrac{1}{2}
  \bigl\|
    \mu_S(x_{t_j};\theta) - \mu_T(x_{t_j})
  \bigr\|_2^{\,2}
\right].
}
\end{equation}
This yields a deterministic specialization of DiffusionOPD in which the student
is trained to match the teacher's one-step transitions directly along its own
rollout trajectory.

\subsection{Discussion: Closed-form KL vs.\ PPO-style Policy Gradient}
\label{sec:discussion_kl_vs_pg}

Our DiffusionOPD objective in Eq.~\eqref{eq:opd-diffusion} already provides a closed-form per-step supervision signal:
\begin{equation}
\mathcal{L}_{\text{OPD}}^{\text{diffusion}}(\theta)
=
\mathbb{E}_{x_{0:N}\sim p_{S,\theta}}
\left[
\sum_{j=0}^{N-1}
\mathrm{KL}\!\Bigl(
p_S(\cdot\mid x_{t_j})\,\Vert\,p_T(\cdot\mid x_{t_j})
\Bigr)
\right],
\end{equation}
with
\begin{equation}
\mathrm{KL}\!\Bigl(
p_S(\cdot\mid x_{t_j})\,\Vert\,p_T(\cdot\mid x_{t_j})
\Bigr)
=
\frac{\|\mu_S(x_{t_j};\theta)-\mu_T(x_{t_j})\|_2^2}{2\bar\sigma_j^{\,2}}.
\label{eq:disc_closed_form_kl}
\end{equation}
Since the student and teacher share the same covariance $\bar\sigma_j^{\,2}I_d$, the KL depends only on the mean mismatch and can be optimized by direct backpropagation.

\subsubsection{Direct Closed-Form KL}
Differentiating Eq.~\eqref{eq:disc_closed_form_kl} gives
\begin{equation}
\nabla_\theta \mathcal{L}_{\text{OPD}}^{\text{diffusion}}(\theta)
=
\mathbb{E}_{x_{0:N}\sim p_{S,\theta}}
\left[
\sum_{j=0}^{N-1}
\frac{\mu_S(x_{t_j};\theta)-\mu_T(x_{t_j})}{\bar\sigma_j^{\,2}}
\cdot
\nabla_\theta \mu_S(x_{t_j};\theta)
\right].
\label{eq:disc_grad_closed_form}
\end{equation}
This is the gradient of the stopped-trajectory surrogate used in practice which does not include derivatives through the student rollout distribution.

\subsubsection{PPO-Style Policy Gradient}
Alternatively, one may regard the teacher model as a process reward model~\cite{wang2024math, zhang2025lessons, cui2025process}, which provides dense per-step supervision along the student trajectory. In this view, a natural choice of per-step advantage is the negative KL,
\[
A_j
=
-\,\mathrm{KL}\!\Bigl(
p_S(\cdot\mid x_{t_j})\,\Vert\,p_T(\cdot\mid x_{t_j})
\Bigr),
\]
and one can optimize a PPO-style surrogate~\cite{schulman2017proximal}:
\begin{equation}
\mathcal{L}_{\text{PG}}(\theta)
=
-\,\mathbb{E}_{a_j\sim \pi_{\theta_{\mathrm{old}}}}
\left[
\min\!\bigl(
\rho_j(\theta)A_j,\,
\mathrm{clip}(\rho_j(\theta),1-\varepsilon,1+\varepsilon)A_j
\bigr)
\right],
\end{equation}
where $\rho_j(\theta)=\pi_\theta(a_j\mid x_{t_j})/\pi_{\theta_{\mathrm{old}}}(a_j\mid x_{t_j})$.

Ignoring clipping, the PPO surrogate reduces to
\begin{equation}
\mathcal{L}_{\text{PG}}(\theta)
=
\,\mathbb{E}_{a_j\sim \pi_{\theta_{\mathrm{old}}}}
\bigl[
\rho_j(\theta)\,\Delta_j(\theta)
\bigr].
\end{equation}
Since the model parameters are held fixed over an entire rollout through gradient accumulation (refer to Algorithm~\ref{alg:DiffusionOPD} for gradient accumulation details), the rollout policy equals the current student policy, i.e., $\pi_{\theta_{\mathrm{old}}}=\pi_\theta$. For a sampled transition, the gradient decomposes as
\begin{equation}
\nabla_\theta \bigl(\rho_j(\theta)\Delta_j(\theta)\bigr)
=
\rho_j(\theta)\nabla_\theta \Delta_j(\theta)
+
\rho_j(\theta)\Delta_j(\theta)\nabla_\theta \log \pi_\theta(a_j\mid x_{t_j}).
\label{eq:pg_integrand_decomp}
\end{equation}
Under $\pi_{\theta_{\mathrm{old}}}=\pi_\theta$, we have $\rho_j(\theta)=1$, so Eq.~\eqref{eq:pg_integrand_decomp} becomes
\begin{equation}
\nabla_\theta \bigl(\rho_j(\theta)\Delta_j(\theta)\bigr)
=
\nabla_\theta \Delta_j(\theta)
+
\Delta_j(\theta)\,\nabla_\theta \log \pi_\theta(a_j\mid x_{t_j}).
\label{eq:disc_pg_decomp}
\end{equation}
where
$\Delta_j(\theta)
:=
\mathrm{KL}\!\Bigl(
p_S(\cdot\mid x_{t_j})\,\Vert\,p_T(\cdot\mid x_{t_j})
\Bigr).$
The first summand in Eq.~\eqref{eq:disc_pg_decomp} is the pathwise
term, and the second is the score-function term.
For a fixed sampled state \(x_{t_j}\) under the detached-rollout surrogate, \(\Delta_j(\theta)\) does not depend on the sampled action \(a_j\). Therefore,
\begin{equation}
\begin{aligned}
\mathbb{E}_{a_j\sim \pi_\theta}
\bigl[
\Delta_j(\theta)\,\nabla_\theta \log \pi_\theta(a_j\mid x_{t_j})
\bigr]
&=
\Delta_j(\theta)\,
\mathbb{E}_{a_j\sim \pi_\theta}
\bigl[
\nabla_\theta \log \pi_\theta(a_j\mid x_{t_j})
\bigr] \\
&=
\Delta_j(\theta)\cdot 0
=0,
\end{aligned}
\end{equation}
Hence the two objectives have the same expected gradient:
\begin{equation}
\mathbb{E}\!\left[\nabla_\theta \mathcal{L}_{\text{PG}}(\theta)\right]
=
\nabla_\theta \mathcal{L}_{\text{OPD}}^{\text{diffusion}}(\theta).
\label{eq:disc_same_expectation}
\end{equation}
Equation~\eqref{eq:disc_same_expectation} shows that direct KL minimization and PPO-style optimization are equivalent \emph{in expectation}. 

\subsubsection{Why the Closed-Form KL Is a Better Solution}
The closed-form KL is preferable to a PPO-style surrogate for two reasons.

First, it yields a lower-variance gradient estimator. The direct objective in Eq.~\eqref{eq:disc_closed_form_kl} is an analytic function of the student transition mean, so its gradient is obtained entirely by pathwise backpropagation. By contrast, the PPO formulation introduces an additional score-function term of the form
$
\Delta_j(\theta)\,
\nabla_\theta \log \pi_\theta(a_j\mid x_{t_j})
$.
For a Gaussian transition with
\begin{equation}
\label{eq:gaussian-action-reparam}
a_j = \mu_S(x_{t_j};\theta)+\bar\sigma_j\epsilon_j,
\qquad \epsilon_j\sim\mathcal N(0,I_d).
\end{equation}
we have
\begin{equation}
\begin{aligned}
\nabla_\theta \log \pi_\theta(a_j\mid x_{t_j})
&=
\frac{\epsilon_j}{\bar\sigma_j}\cdot \nabla_\theta \mu_S(x_{t_j};\theta).
\end{aligned}
\end{equation}
Thus, the PPO estimator contains an additional stochastic term proportional to Gaussian noise. Although this term is unbiased in expectation, it introduces nonzero gradient variance, which is absent in the closed-form KL objective.

Second, the closed-form KL loss formulation remains valid in both stochastic and deterministic sampling regimes. In the deterministic ODE regime, we can use Eq.~\eqref{eq:opd-ode} to update student policy. 
A PPO-style objective, however, is inherently tied to a stochastic policy density through $\log \pi_\theta$ and the importance ratio $\rho_j$. 

Therefore, for DiffusionOPD, the closed-form KL is not only lower-variance but also applicable to a wider range of samplers, covering both SDE and ODE samplers within a single training principle.

\subsection{Training Recipe}
\begin{algorithm}[t]
\caption{DiffusionOPD}
\label{alg:DiffusionOPD}
\small
\KwIn{Tasks $\mathcal{M}=\{1,\dots,M\}$; prompt datasets $\{\mathcal{C}^{(m)}\}_{m=1}^{M}$; pretrained diffusion policy $\mathbf{v}^{\mathrm{ref}}$; denoising schedule $\{t_j,\bar{\sigma}_j^2\}_{j=0}^{N-1}$.}
\KwOut{Unified multi-task diffusion student $\mathbf{v}_\theta$.}

\textbf{Stage 1: Per-task teacher training}\;
\ForEach{$m \in \mathcal{M}$}{
    Train a task-specific teacher $\mathbf{v}_{\phi_m}^{(m)}$ for task $m$ using an off-the-shelf RL algorithm\;
}

\textbf{Stage 2: Multi-task on-policy distillation}\;
Initialize $\mathbf{v}_\theta \leftarrow \mathbf{v}^{\mathrm{ref}}$\;

\For{each training round}{
    Initialize total round loss $\mathcal{L}_{\text{total}} \leftarrow 0$\;
    \For{$m = 1, \dots, M$}{
        Sample prompts $\vc \sim \mathcal{C}^{(m)}$\;
        Roll out the current student $\mathbf{v}_\theta$ on $\vc$ to obtain on-policy trajectory $\{\vx_{t_j}\}_{j=0}^{N}$ \tcp*{\texttt{no\_grad}}
        Compute task loss $\mathcal{L}_m$ via Monte Carlo estimate of Eq.~\eqref{eq:opd-diffusion} or Eq.~\eqref{eq:opd-ode} using teacher $\mathbf{v}_{\phi_m}^{(m)}$ \tcp*{Eq.~\eqref{eq:opd-ode} by default}
        $\mathcal{L}_{\text{total}} \mathrel{{+}{=}} \mathcal{L}_m$
    }
    Update $\theta$ by performing one backward pass on $\mathcal{L}_{\text{total}}$ and taking an optimizer step\;
}
\end{algorithm}

Our DiffusionOPD follows a two-stage training paradigm, as summarized in Algorithm~\ref{alg:DiffusionOPD}. In the first stage, we decompose the multi-task problem into \(M\) individual tasks and train a separate task-specific teacher \(\mathbf{v}_{\phi_m}^{(m)}\) for each task \(m \in \mathcal{M}\) using off-the-shelf diffusion RL algorithms~\cite{zheng2025diffusionnft, wang2025grpo}. This stage allows each teacher to specialize in its own reward objective without being affected by inter-task interference.

In the second stage, we distill these specialized teachers into a single unified student \(\mathbf{v}_\theta\), initialized from the pretrained diffusion policy \(\mathbf{v}^{\mathrm{ref}}\). Training proceeds in a round-robin on-policy manner over all tasks. For each task \(m\), we first sample prompts from \(\mathcal{C}^{(m)}\), then roll out the current student to obtain an on-policy denoising trajectory \(\{ \vx_{t_j} \}_{j=0}^{N}\). Along this sampled trajectory, we evaluate the corresponding task teacher and compute a Monte Carlo estimate of the OPD objective in Eq.~\eqref{eq:opd-diffusion}, which matches the student and teacher transition means at every denoising step.

To stabilize multi-task optimization, we accumulate losses over a full round-robin cycle before updating the student. Concretely, we set the gradient accumulation factor to \(G=M\), i.e., one accumulation step per task, and average the task losses within each round. A single backward pass and optimizer step are performed only after all \(M\) tasks have been visited once. This design makes each parameter update reflect the supervision from the complete task set, reducing update variance and mitigating bias toward any individual task.

\section{Experiments}
\label{sec:experiments}

\begin{table*}[t]
    \centering
    \caption{\textbf{Evaluation Results.} \hl{Gray-colored}: in-domain reward. $^\ddagger$Evaluated at 1024$\times$1024 resolution. \textbf{Bold}: best; \underline{Underline}: second best. $^\ast$Approximated training time. Wall-clock time is reported in hours. The \textbf{Average} column denotes the mean of min-max normalized scores over all metrics. $^\dagger$For DiffusionOPD, wall-clock time is reported as the maximum teacher training time plus OPD training time.}
    \resizebox{\linewidth}{!}{
        \begin{tabular}{lcccccccccc}
            \toprule
            \multirow{2}{*}{\textbf{Model}} & \multirow{2}{*}{\textbf{Wall-clock time}} & \multicolumn{2}{c}{\textbf{Rule-Based}} & \multicolumn{6}{c}{\textbf{Model-Based}} & \multirow{2}{*}{\textbf{Average}} \\ 
            \cmidrule(lr){3-4} \cmidrule(r){5-10}
            && \textbf{GenEval} & \textbf{OCR} & \textbf{PickScore} & \textbf{ClipScore} & \textbf{HPSv2.1} & \textbf{Aesthetic} & \textbf{ImgRwd} & \textbf{UniRwd} & \\ 
            \midrule
            SD-XL$^\ddagger$ & — & 0.55 & 0.14 & 22.42 & 0.287 & 0.280 & 5.60 & 0.76 & 2.93 & 0.390 \\
            SD3.5-L$^\ddagger$ & — & 0.71 & 0.68 & 22.91 & 0.289 & 0.288 & 5.50 & 0.96 & 3.25 & 0.601 \\
            FLUX.1-Dev & — & 0.66 & 0.59 & 22.84 & 0.295 & 0.274 & 5.71 & 0.96 & 3.27 & 0.599 \\
            \midrule
            SD3.5-M (w/o CFG) & — & 0.24 & 0.12 & 20.51 & 0.237 & 0.204 & 5.13 & -0.58 & 2.02 & 0.000 \\
            + CFG & — & 0.63 & 0.59 & 22.34 & 0.285 & 0.279 & 5.36 & 0.85 & 3.03 & 0.484 \\
            GenEval Teacher & 46.92 & \cellcolor{llgray}\textbf{0.96} & 0.40 & 22.04 & 0.274 & 0.248 & 5.24 & 0.59 & 2.97 & 0.473 \\
            OCR Teacher & 33.17 & 0.65 & \cellcolor{llgray}0.93 & 22.27 & 0.290 & 0.272 & 5.26 & 0.90 & 3.09 & 0.550 \\
            Aes Teacher & 85.75 & 0.49 & 0.59 & \cellcolor{llgray}\textbf{24.02} & \cellcolor{llgray}0.295 & \cellcolor{llgray}\textbf{0.346} & \textbf{6.22} & \underline{1.498} & 3.48 & 0.698 \\
            \midrule
            Multi-Task GRPO-Guard & 129.86 & \cellcolor{llgray}0.89 & \cellcolor{llgray}\underline{0.94} & \cellcolor{llgray}23.12 & \cellcolor{llgray}\underline{0.296} & \cellcolor{llgray}0.307 & 5.61 & 1.31 & 3.33 & 0.763 \\
            Multi-Task NFT & 128.42 & \cellcolor{llgray}\underline{0.95} & \cellcolor{llgray}\textbf{0.96} & \cellcolor{llgray}22.59 & \cellcolor{llgray}0.288 & \cellcolor{llgray}0.282 & 5.41 & 1.08 & 3.23 & 0.715 \\
            Cascade NFT & 148.49$^\ast$ & \cellcolor{llgray}0.94 & \cellcolor{llgray}0.91 & \cellcolor{llgray}23.80 & \cellcolor{llgray}0.293 & \cellcolor{llgray}0.331 & 6.01 & 1.49 & \underline{3.49} & \underline{0.851} \\
            \textbf{DiffusionOPD(Ours)} & 85.75+11.26$^\dagger$ & \cellcolor{llgray}\textbf{0.96} & \cellcolor{llgray}\underline{0.94} & \cellcolor{llgray}\underline{23.99} & \cellcolor{llgray}\textbf{0.297} & \cellcolor{llgray}\underline{0.342} & \underline{6.15} & \textbf{1.50} & \textbf{3.50} & \textbf{0.929} \\
            \bottomrule
        \end{tabular}
    }
    \label{tab:all_task}
\end{table*}

In this section, we detail the experimental setup and demonstrate the capabilities of DiffusionOPD from three perspectives: (1) comparison with major multi-task learning baselines, (2) comparison with alternative distillation methods for transferring knowledge from multiple single-task teachers, and (3) ablation studies on key design choices.

\subsection{Experimental Setup}

\subsubsection{Implementation Details}
We follow DiffusionNFT~\cite{zheng2025diffusionnft} for the experimental setup and use SD3.5-Medium~\cite{esser2024scaling} at 512$\times$512 resolution as the base model. 
Our reward models include both rule-based and model-based signals. The rule-based rewards are GenEval~\cite{ghosh2023geneval} for compositional generation and OCR for visual text rendering, while the model-based rewards include PickScore~\cite{kirstain2023pick}, ClipScore~\cite{hessel2021clipscore}, HPSv2.1~\cite{wu2023human}, Aesthetics~\cite{schuhmann2022aesthetics}, ImageReward~\cite{xu2023imagereward}, and UnifiedReward~\cite{wang2025unified}. 
For data, we use the FlowGRPO splits for GenEval and OCR, and train on Pick-a-Pic~\cite{kirstain2023pick} while evaluating on DrawBench~\cite{saharia2022photorealistic} for the model-based rewards. We also adopt the same finetuning and evaluation configuration as DiffusionNFT, using LoRA ($\alpha=64$, $r=32$) and a 40-step first-order ODE sampler for evaluation.

\subsubsection{Single-Task Teachers}
We select the training algorithm for each teacher according to the characteristics of its reward task. For OCR and Aesthetics, we train the teachers with GRPO-Guard. In our preliminary experiments, although DiffusionNFT converges rapidly, it is highly susceptible to reward hacking on OCR, often achieving high reward scores at the cost of severe image quality degradation. For the aesthetics teacher, we optimize an equally weighted (1:1:1) mixture of \texttt{PickScore}, \texttt{ClipScore}, and \texttt{HPSv2.1}, and find that GRPO-Guard consistently attains a higher performance ceiling than DiffusionNFT on this objective. For GenEval, we instead use DiffusionNFT to train the teacher, as it exhibits faster convergence and a higher performance ceiling on this task.

\subsubsection{Baselines}
We compare DiffusionOPD against several competitive baselines: (1) \textbf{Single-task teachers}, i.e., the specialized models described above; (2) \textbf{Multi-Task RL}, which uses different RL algorithms to jointly train on multiple tasks by alternating across the corresponding datasets in the same curriculum as DiffusionOPD; and (3) \textbf{Cascade NFT}~\cite{zheng2025diffusionnft}, a sequential training baseline where different tasks are learned stage by stage.

\subsection{Comparisons with Multi-Task RL Methods}
\label{main_results}

\begin{figure*}
    \centering
    \includegraphics[width=\linewidth]{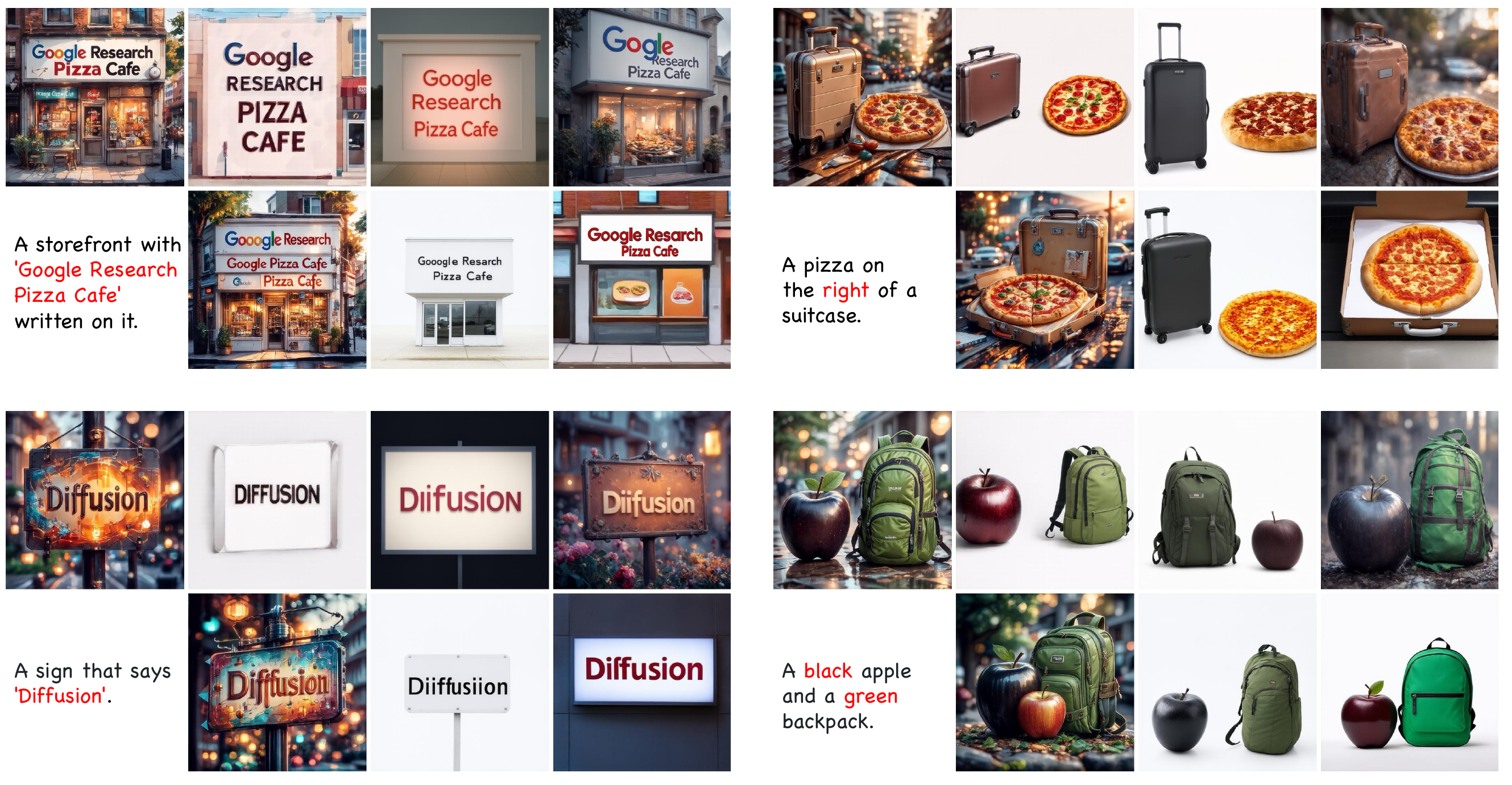}
    \caption{Qualitative comparisons against multi-task RL methods and single-task teachers. Each case is presented in two rows. The first row shows, from left to right, DiffusionOPD (ours), Multi-Task GRPO-Guard, Multi-Task NFT, and Cascade NFT. The second row shows the input prompt, our Aes Teacher, our GenEval Teacher, and our OCR Teacher.}
    \label{fig:comparison_rl}
\end{figure*}

Table~\ref{tab:all_task} shows that single-task teachers are highly specialized to their own training domains, but generalize poorly across heterogeneous rewards. The GenEval Teacher mainly excels at compositional alignment, the OCR Teacher is strongest on text rendering, and the Aes Teacher performs best on aesthetic-related objectives, while each of them shows limited transferability beyond its own optimization target. 
Multi-task RL methods improve overall task coverage, but require substantially longer training time and still struggle on more challenging objectives such as aesthetics, indicating slower convergence and stronger optimization interference across domains. 
Although Cascade NFT achieves relatively competitive performance, it is the slowest and most cumbersome strategy due to sequential multi-stage training, and is also prone to catastrophic forgetting, which limits its final performance. 

\begin{figure*}
    \centering
    \includegraphics[width=\linewidth,height=0.82\textheight,keepaspectratio]{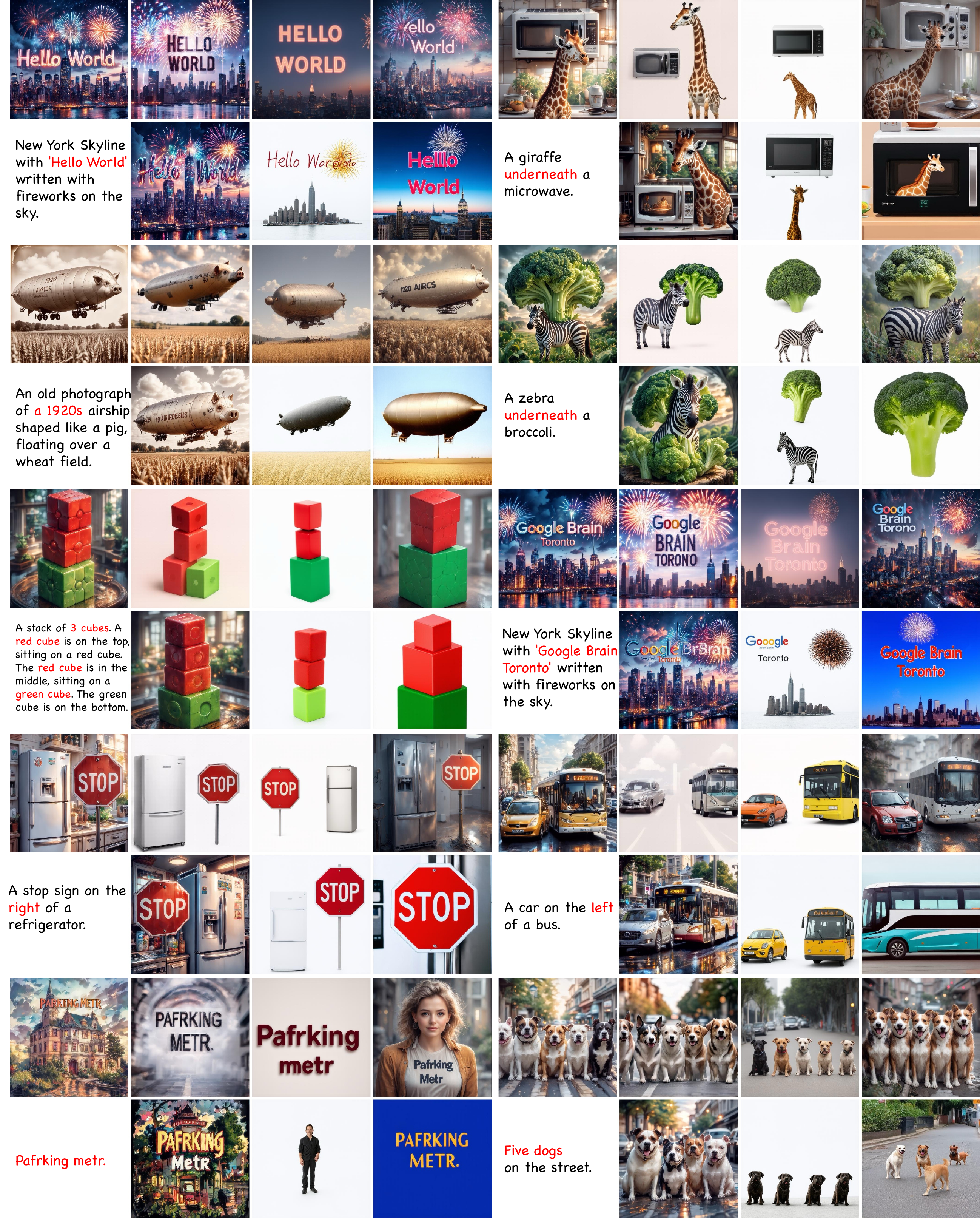}
    \caption{Qualitative comparisons against multi-task RL methods and single-task teachers. Each case is presented in two rows. The first row shows, from left to right, DiffusionOPD (ours), Multi-Task GRPO-Guard, Multi-Task NFT, and Cascade NFT. The second row shows the input prompt, our Aes Teacher, our GenEval Teacher, and our OCR Teacher.}
    \label{fig:comparison_rl_more}
\end{figure*}

By contrast, DiffusionOPD achieves the best overall performance, demonstrating the effectiveness of our training paradigm for multi-domain preference optimization. Qualitative comparison in Figure~\ref{fig:comparison_rl} and ~\ref{fig:comparison_rl_more} also demonstrates the superior visual quality of our method.

\begin{figure*}
    \centering
    \includegraphics[width=\linewidth]{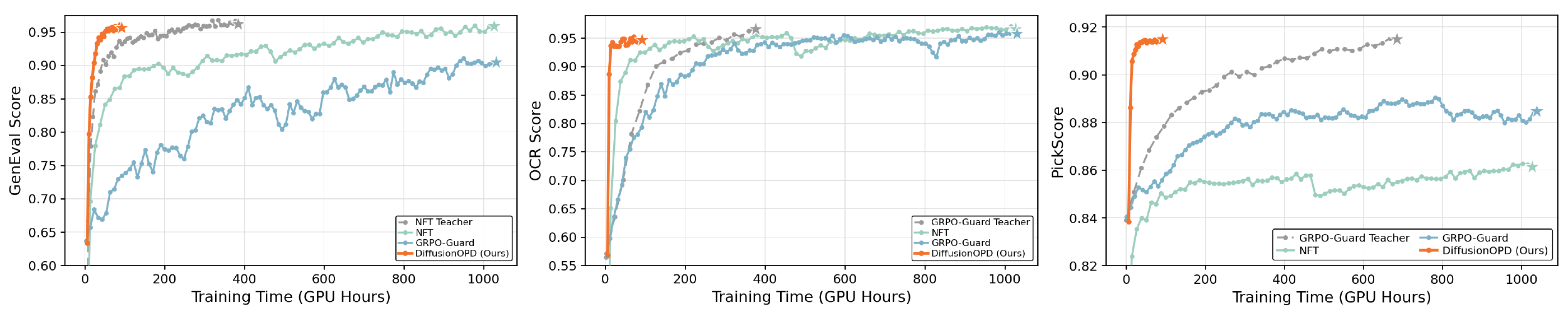}
    \caption{DiffusionOPD outperforms multi-task RL baselines in both efficiency and performance.}
    \label{fig:main_curve}
\end{figure*}

To further evaluate training efficiency, we plot the convergence curves in Figure~\ref{fig:main_curve}. As shown, multi-task RL baselines converge more slowly than single-task RL teachers, indicating that jointly optimizing heterogeneous rewards introduces severe optimization interference and hinders learning efficiency.
Besides, DiffusionOPD requires much less total training time than the Multi-Task RL baselines to reach the same target score, while also attaining a substantially higher performance ceiling.

\subsection{Ablation Studies}
\label{ablation}
\subsubsection{Distillation Methods}
\begin{figure}
    \centering
    \includegraphics[width=\linewidth]{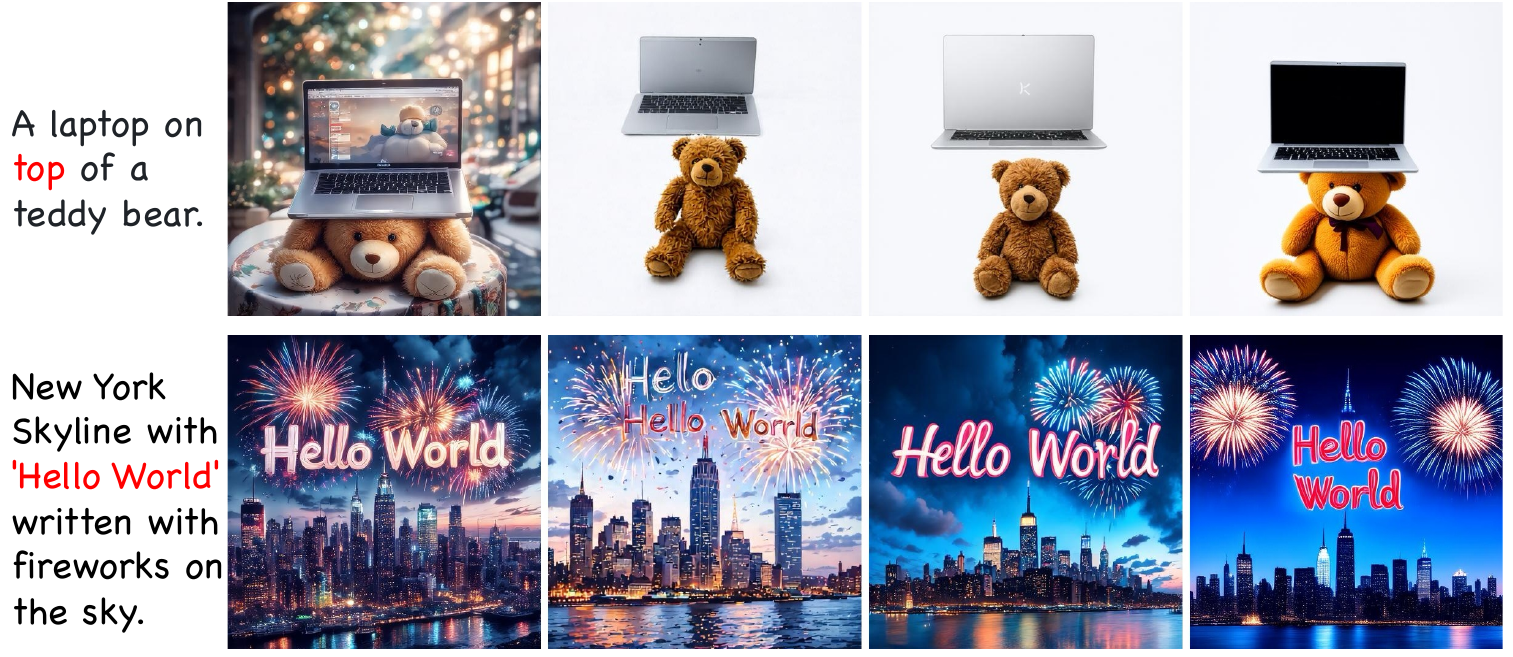}
    \caption{Qualitative comparisons with different distillation methods. From left to right: DiffusionOPD (ours), DMD, TDM, and SFT.}
    \label{fig:comparison_distill}
\end{figure}

\begin{figure}
    \centering
    \includegraphics[width=\linewidth]{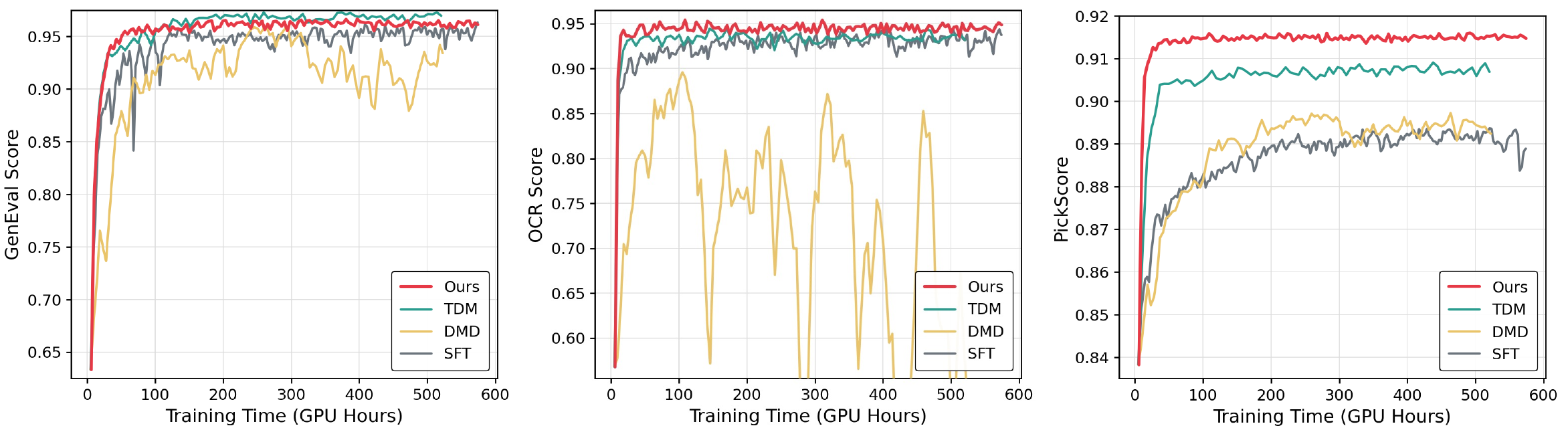}
    \caption{Ablation studies on distillation methods. SFT is trained on images generated from teacher rollouts, while all other baselines use student on-policy rollouts and distill from the same set of teacher models using their respective objectives.}
    \label{fig:additional_comparison}
\end{figure}

\begin{figure*}
    \centering
    \includegraphics[width=0.7\linewidth]{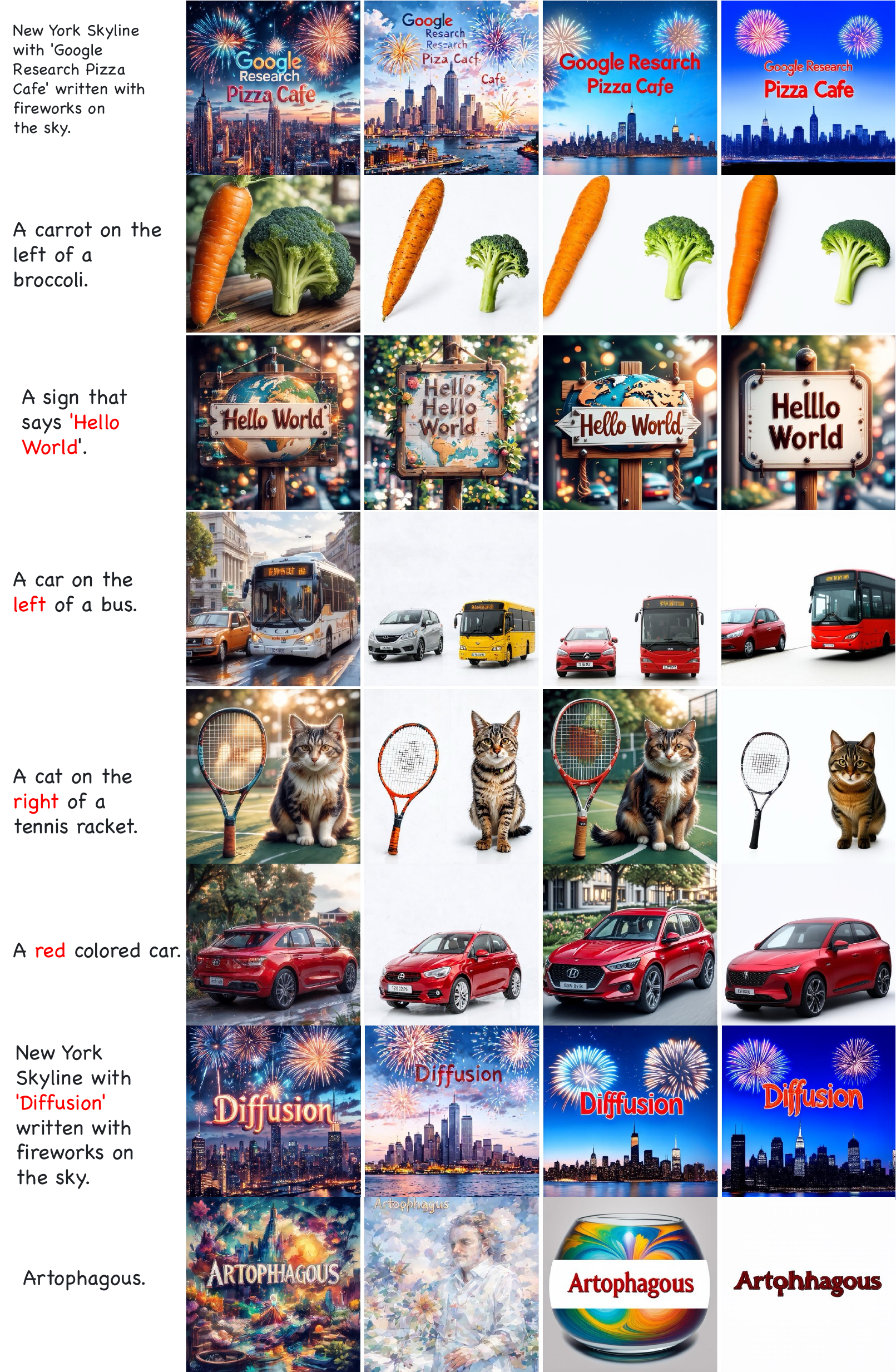}
    \caption{Qualitative comparisons with different distillation methods. From left to right: DiffusionOPD (ours), DMD, TDM, and SFT.}
    \label{fig:comparison_distill_more}
\end{figure*}

We further compare DiffusionOPD with several representative distillation baselines that transfer knowledge from single-task teachers, including DMD~\cite{yin2024improved}, TDM~\cite{luo2025learning}, and supervised fine-tuning (SFT).
For SFT, we use the corresponding teacher to generate images online and train the student to imitate these teacher-generated samples, which can also be viewed as a form of teacher knowledge distillation.
For DMD and TDM, we perform on-policy sampling using the student model and distill the corresponding teacher through the training gradients defined by each method.
To ensure a fair comparison, we implement all baselines under the same setting as DiffusionOPD: each method is distilled from the identical set of specialized teachers, and training is conducted by alternating across datasets.
As shown in Figure~\ref{fig:additional_comparison}, DiffusionOPD consistently achieves the fastest convergence and highest final performance ceiling among all compared distillation methods. Qualitative results in Fig.~\ref{fig:comparison_distill} and Fig.~\ref{fig:comparison_distill_more} also demonstrate our superiority.

\subsubsection{Loss Formulation}
To validate our analysis in Section~\ref{sec:discussion_kl_vs_pg}, we compare the closed-form KL objective against the PPO-style policy gradient. To ensure a fair comparison, both methods are evaluated in the multi-task setting with an identical sampling noise level of $a=0.7$.
As shown in Figure~\ref{fig:Ablation_sampler}, under the same noise level, the closed-form KL objective achieves faster reward improvement and a higher performance ceiling than PPO-style policy gradients.

\subsubsection{Noise Level}
\begin{figure}
    \centering
    \includegraphics[width=\linewidth]{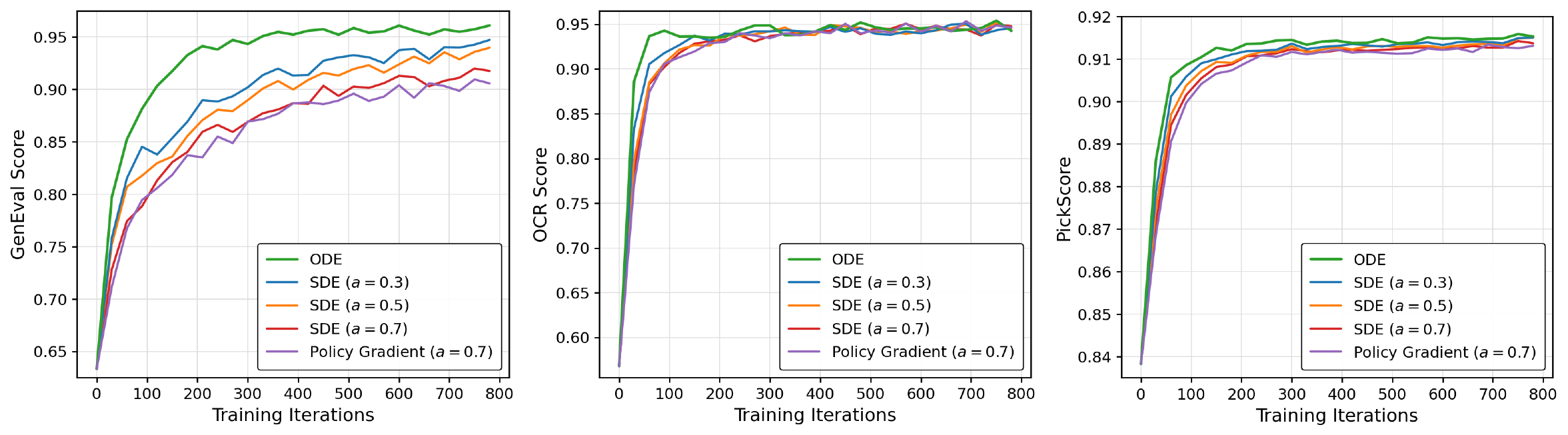}
    \caption{Ablation studies on the loss formulation and sampler noise level. When the noise level is set to $0$, the SDE sampler reduces to an ODE sampler, and the student is optimized using Eq.~\eqref{eq:opd-ode}. As shown, the PPO-style policy gradient underperforms its closed-form KL counterpart. Moreover, lower noise levels lead to faster convergence and higher performance ceiling.}
    \label{fig:Ablation_sampler}
\end{figure}

We further conduct an ablation study on the noise level of the SDE sampler used during distillation. As shown in Figure~\ref{fig:Ablation_sampler}, reducing the noise level consistently leads to faster convergence and higher evaluation scores for the student model. In particular, the ODE sampler with is up to five times more efficient than the SDE sampler with noise level=0.7.

\section{Conclusion}
We introduced \textbf{DiffusionOPD}, a new on-policy distillation paradigm for multi-task training of diffusion models. By decoupling single-task exploration from multi-task capability integration, DiffusionOPD avoids the optimization conflict of joint multi-task RL and the inefficiency and forgetting of cascade RL.
We further developed a principled theoretical framework that extends OPD to diffusion Markov chains, yielding a closed-form per-step reverse-KL objective that unifies stochastic SDE and deterministic ODE refinement. Compared with PPO-style policy-gradient optimization, this objective enables lower-variance training and applies naturally across sampler types.
Extensive experiments and ablations show that DiffusionOPD consistently improves both training efficiency and final performance over prior baselines, achieving state-of-the-art results on aesthetics, OCR, and GenEval. We hope DiffusionOPD can serve as a useful foundation for future work on multi-task and preference-aligned diffusion modeling.

\begin{acks}
This work is supported by the National Natural Science Foundation of China (Grant No. 62521004) and the Science and Technology Commission of Shanghai Municipality (No. 25511106100).
\end{acks}

\clearpage

\clearpage

\bibliographystyle{ACM-Reference-Format}
\bibliography{main}

\end{document}